\pdfoutput=1

\documentclass[11pt]{article}
\usepackage{amsfonts}
\usepackage[preprint]{latex/acl}

\usepackage{times}
\usepackage{latexsym}
\usepackage{amsmath}
\usepackage{cleveref}
\usepackage[table]{xcolor}
\definecolor{lightyellow}{RGB}{255,249,220}
\usepackage{xcolor}
\definecolor{TiffanyBlue}{RGB}{10,186,181}
\usepackage[most]{tcolorbox}
\usepackage{amssymb}

\usepackage[T1]{fontenc}

\usepackage[utf8]{inputenc}

\usepackage{microtype}
\usepackage{booktabs}
\usepackage{multirow}

\usepackage{inconsolata}

\usepackage{graphicx}

\newcommand{\ourwork}{AtomMem~}

\title{\ourwork: Learnable Dynamic Agentic Memory \\with Atomic Memory Operation}

\author{Yupeng Huo\textsuperscript{1},
Yaxi Lu\textsuperscript{2},
Zhong Zhang\textsuperscript{2},
Haotian Chen\textsuperscript{2},
Yankai Lin\textsuperscript{1}\thanks{Corresponding author}
\\
\textsuperscript{1} Renmin University of China,
\textsuperscript{2} Tsinghua University \\
\texttt{\{huoyupeng, yankailin\}@ruc.edu.cn} \\
\textbf{Project: \url{https://github.com/RUCBM/AtomMem}}
}

\begin{document}
\maketitle
\begin{abstract}
Equipping agents with memory is essential for solving real-world long-horizon problems. However, most existing agent memory mechanisms rely on static and hand-crafted workflows. This limits the performance and generalization ability of these memory designs, which highlights the need for a more flexible, learning-based memory framework. In this paper, we propose \ourwork, which reframes memory management as a dynamic decision-making problem. We deconstruct high-level memory processes into \textbf{fundamental atomic CRUD} (Create, Read, Update, Delete) operations, transforming the memory workflow into a learnable decision process. Through widely used reinforcement learning~(GRPO), \ourwork learns an autonomous, task-aligned policy to orchestrate memory behaviors tailored to specific task demands. Experimental results across 3 long-context QA benchmarks and 2 web benchmarks demonstrate that the trained model consistently outperforms prior static-workflow memory methods. 
Further analysis of training dynamics shows that our learning-based formulation enables the agent to discover structured, task-aligned memory management strategies, highlighting a key advantage over predefined workflows.
\end{abstract}

\section{Introduction}
Enabling LLM-based agents to accomplish long-horizon and more complex tasks has been a shared goal across both industry and academia~\citep{Chen2025ReinforcementLF,pmlr-v267-erdogan25a,wang2025ragenunderstandingselfevolutionllm}. A critical bottleneck in this pursuit is the design of memory mechanisms. Currently, most memory mechanisms of LLM-based agents rely on static, expert-crafted workflows~\citep{Amem,mem0,memos}. In these systems, memory operations are confined to predefined pipelines rather than decided autonomously by the model.

\begin{figure}
    \centering
    \includegraphics[width=\linewidth]{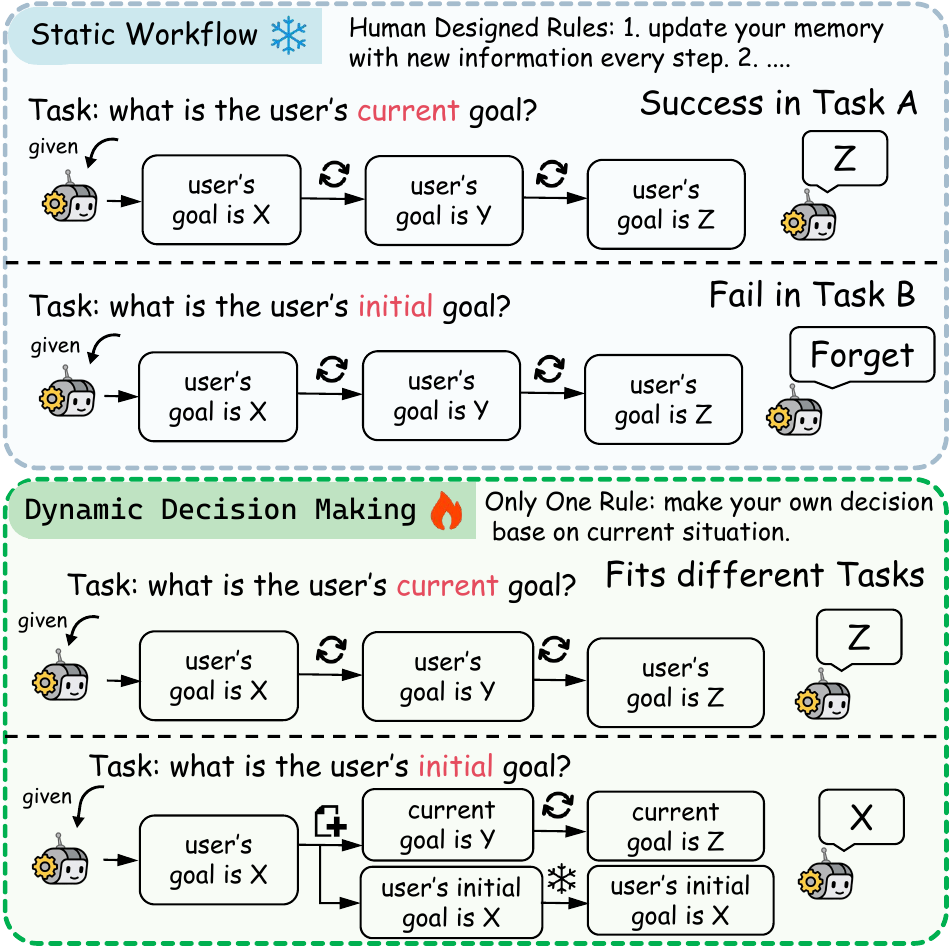}
    \caption{The one-size-fits-all workflow of static memory often fails to adapt to diverse tasks. Instead, a dynamic memory system is needed to determine the optimal memory strategy based on the specific situation.}
    \label{fig:motivation}
\end{figure}

A core limitation of these approaches lies in their implicit `one-size-fits-all' assumption: they impose fixed memory management rules for information retention, rather than allowing the model to make autonomous decisions. 
Strategies like continuous memory fusion~\citep{Amem} or predefined forgetting schedules~\citep{MemoryBank} may work well in generic scenarios but fail in complex environments. For example, exponential forgetting schedules may prematurely discard early yet critical cues in long-horizon reasoning~\citep{MemoryLLM, M+}. As illustrated in \Cref{fig:motivation}, the same static workflow successfully preserves important information in Task A, but fails to do so in Task B. This naturally raises a question: how should we design a more adaptive and effective agent memory system?


To answer this question, we propose \ourwork, which reframes memory management of LLM-based agents not as a fixed workflow, but as a decision-making problem. Drawing inspiration from agent tool learning~\citep{toolLLM,llmtoolssurvey}—where models learn when to invoke tools based on context, we deconstruct high-level memory processes into their fundamental atoms: the standard CRUD (Create, Read, Update, Delete) operations. 
This atomization transforms a static memory workflow into a learnable decision process~\citep{semi-MDPs,HRL}.

A key advantage of this framework is that the effectiveness of memory management is no longer fundamentally bounded by expert rules, but instead by the model's own capacity to make appropriate decisions. By training with reinforcement learning (RL), the agent can acquire experience over these atomic operations and gradually learn a policy for managing memory in a task-aware manner—retaining information that is important for the completion of the task. In this way, memory is no longer treated as a fixed mechanism, but as a kind of policy behavior optimized through interaction with the environment.

We evaluate our method on multiple memory-intensive tasks, which can be broadly divided into two categories:
1) Three multi-hop long-context QA tasks, including HotpotQA~\citep{hotpotQA}, 2WikiMultihopQA~\citep{2wiki}, Musique~\citep{Musique}, and 2) Two multi-turn web search task including GAIA~\citep{GAIA} and WebWalkerQA~\citep{WebWalker}. Across all tasks, our approach consistently outperforms prior methods reliant on static memory workflows by approximately 3-8 percentage points under the same Qwen3-8B backbone. 
These results demonstrate that treating memory management as an atomic-level capability optimized via RL is more effective than relying on predefined routines.

Beyond overall performance gains, we further uncover an empirical insight into how memory should be managed for these tasks. 
The learned policy exhibits a systematic shift in memory operation usage: the frequencies of Create, Update, and Delete operations steadily increase, while reliance on Read actions decreases and stabilizes at a lower level. 
Whereas when the task condition changes, the frequencies follow entirely different trends.
This suggests that effective memory control for these tasks benefits from learning task-aligned patterns, rather than maintaining a fixed strategy.


\begin{figure*}
    \centering
    \includegraphics[width=\linewidth]{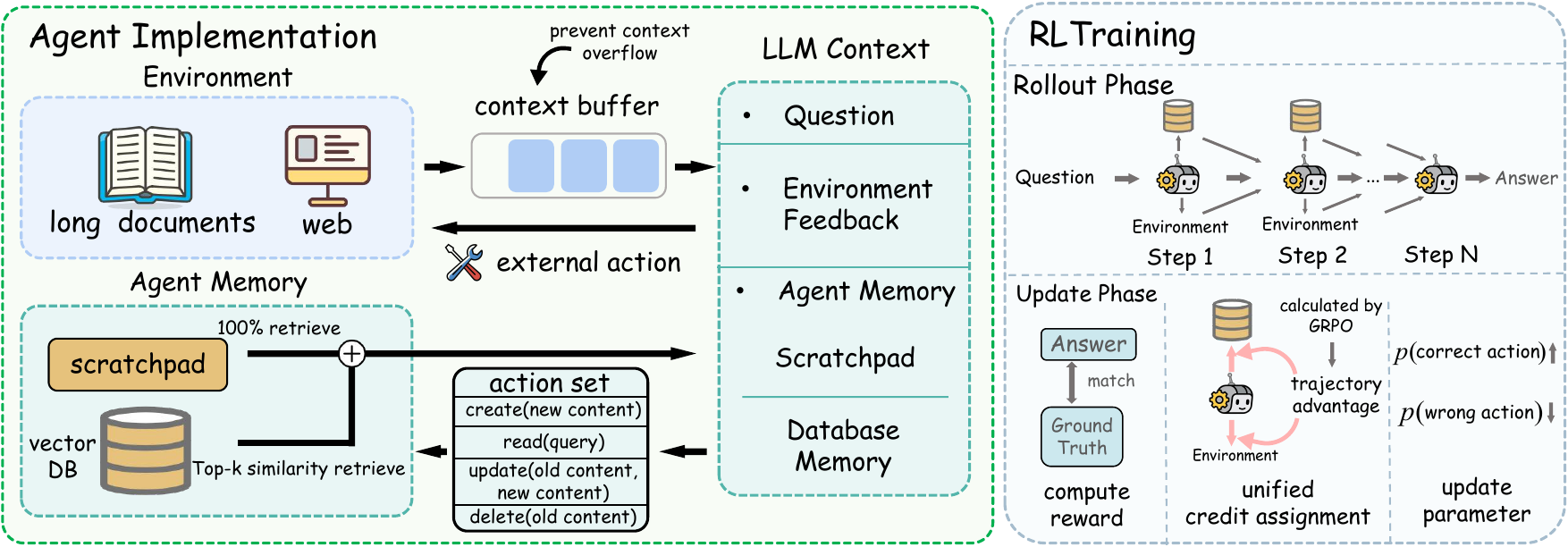}
    \caption{Overview of the \ourwork framework. The agent interacts with environments while maintaining an external memory. High-level memory workflows are decomposed into atomic CRUD (Create, Read, Update, Delete) operations. Through end-to-end reinforcement learning, the agent learns a task-aligned memory management policy that dynamically decides when to store, retrieve, update, or delete information based on task demands.}
    \label{fig:main}
\end{figure*}


\section{Related Works}
\paragraph{Static Memory Workflow}
Early memory mechanisms in LLM-based agents typically relied on heuristic-based static workflows. These memory mechanisms can be categorized into two types: 1) \textbf{Imitation-Based}: Imitation-based approaches refer to transferring designs from natural systems or other engineering domains into agent memory architectures. For example, MemoryBank~\citep{MemoryBank} draws an analogy between agent memory and human memory, while MemGPT~\citep{MemGPT} likens the agent’s context to computer memory.  
2) \textbf{Prior-Based}: Prior-based approaches refer to carefully crafted workflows designed by human experts based on prior knowledge~\citep{MemTree,ChatDB,MemoRAG,SCM}.
Despite their theoretical appeal, these methods share a common limitation: the memory workflow is hard-coded by experts. 
This rigidity prevents the agent from adapting its memory strategy to different tasks. For example, architectures designed for QA tasks may be difficult to transfer to agent tasks that interact with the external environment. In contrast, our work moves beyond static rules, aiming to learn a dynamic memory policy directly from different task.


\paragraph{Reinforcement Learning in Agent Memory}
As reinforcement learning becomes a common method for fine-tuning LLM behavior~\citep{DAPO,RAGEN}, some works have begun using RL to enhance agent memory, which we categorize into two paradigms:
\textbf{Summarization-Based}: MemAgent~\citep{MemAgent} and Mem1~\citep{MEM1} utilize step-wise overwriting summaries. Although overwriting can theoretically emulate any atomic operation, the workflow is restricted to a mandatory ``update-at-every-step'' routine. This ignores information density, forcing redundant updates even when new data is sparse. 
\textbf{Heuristic-Tool-Based}: Memory-As-Action~\citep{memory-as-action} and AgentFold~\citep{AgentFold} introduce memory management tools such as context pruning or folding. While providing some dynamic control over memory, the tool designs themselves rely heavily on manual priors. In contrast, we provids the model with only the most atomic memory operations, better highlighting the characteristics of the memory-as-decision-making paradigm.

\section{Method}
In this section, we formulate the memory management of LLM-based agents as a sequential decision-making problem and introduce a complete action space over memory operations.

\subsection{Preliminaries: POMDP for Memory}
We model memory management in an LLM-based agent as a Partially Observable Markov Decision Process (POMDP) $(\mathcal{S}, \mathcal{A}, P, \Omega, \mathcal{O}, \mathcal{R}, \gamma)$, where memory is explicitly treated as a controllable component of the environment.

\noindent $\triangleright$ \textbf{Global State $\mathcal{S}$}. 
The global state is $s_t = (s_t^{env}, s_t^{mem})$, comprising the external environment and the internal memory state.

\noindent $\triangleright$ \textbf{Action Space $\mathcal{A}$}. An action $a_t \in \mathcal{A}$ is a joint decision $a_t = (a_t^{env}, a_t^{mem})$. While $a_t^{env}$ represents task-specific execution (e.g., search), $a_t^{mem}$ denotes a memory management action chosen from our atomic CRUD space.

\noindent $\triangleright$ \textbf{Transition Function $P$}. The transition $P(s_{t+1} | s_t, a_t)$ defines how the state evolves. Notably, the internal memory state $s_{t+1}^{mem}$ is directly modified by the agent's actions.

\noindent $\triangleright$ \textbf{Observation Function $\mathcal{O}$}.
The agent observes $o_t = (o_t^{env}, o_t^{mem})$. Crucially, memory is not fully observable: $o_t^{mem}$ is determined by previous memory actions (e.g., Read), making memory access an explicit decision variable. 

When we formalize memory as a POMDP, it means that we can leverage RL algorithms to optimize the LLM's ability to manage memory.
Notably, the agent’s memory should be regarded as part of the environment and reset at the start of each independent task, meaning that
\begin{equation}
    s_0^{mem} = \varnothing
\end{equation}
This setting should be distinguished from another type of memory that accumulates experience across different tasks~\citep{Expel, Memp}.

\subsection{Why Atomic CRUD Operations?}
We adopt CRUD (Create, Read, Update, Delete) as the atomic memory action space $\mathcal{A}$ based on three foundational properties.

\noindent $\triangleright$ \textbf{Completeness.}
CRUD constitutes a universal set of operations capable of synthesizing state-transition in memory. Any valid memory state can be reached from the current state through a series of CRUD operations. This guarantees that, under ideal optimization, the agent's memory can achieve its maximal potential performance.

\noindent $\triangleright$ \textbf{Atomic Minimality.}
Any higher-level memory tool design can be viewed as invoking a structured combination of CRUD operations. In contrast, the invocation of CRUD primitives themselves cannot be further decomposed into other finer-grained tool designs. 

\noindent $\triangleright$ \textbf{Task-agnosticness}
The CRUD action framework is not tied to any specific downstream task. Its ability to complete tasks relies entirely on the LLM’s decision-making capability at a given step, which is itself optimizable. Consequently, this action set constitutes a potential foundation for a general-purpose agent memory.

In summary, rather than proposing isolated memory tools, we focus on a complete and general-purpose operator set that governs memory state evolution.

\subsection{Memory Mechanism Implementation}

We model memory at step $t$ as a dynamic set
\begin{equation}
    \mathcal{M}_t = \{ m_i \}_{i=1}^{N_t},
\end{equation}
where $m_i$ encodes a stored memory entry.

Memory manipulation is exposed as a learnable action space
\begin{equation}
    \mathcal{A}^{mem} = \{\text{Create}, \text{Read}, \text{Update}, \text{Delete}\},
\end{equation}
where each primitive defines a state transition operator over $\mathcal{M}_t$.

At each decision step $t$, conditioned on observation $o_t$, the policy generates a sequence of memory actions
\begin{equation}
    \mathcal{A}_t = \{ a_t^{1}, \ldots, a_t^{K_t} \},
\end{equation}
which forms a compositional macro-action within a single environment step.
Non-read operations are executed sequentially, yielding a composed transition
\begin{equation}
    \mathcal{M}_{t+1} = a_t^{K_t} \circ \cdots \circ a_t^{1}(\mathcal{M}_t),
\end{equation}
where $a_t^{k} \in \{\text{Create}, \text{Update}, \text{Delete}\}$.

\paragraph{Hybrid Memory Retrieval}
The Read operation does not alter the memory state. Instead, it retrieves information from $\mathcal{M}_t$ and produces a memory observation: content requested at step $t-1$ forms the observation at step $t$.
We implement a hybrid retrieval mechanism that combines deterministic retrieval with selective query-based retrieval:

\noindent $\triangleright$ \textbf{Deterministic Retrieval (Scratchpad).}
A special memory entry $m_t^{scr}$ is retrieved at every step. This scratchpad captures the global task state and preserves pivotal information necessary for step-wise decision-making. Functionally, it is identical to other memory entries, differing only in its mandatory retrieval schedule.

\noindent $\triangleright$ \textbf{Selective Retrieval.}
The agent generates a textual query $q_t$ as a tool parameter, and relevant memory entries are retrieved based on semantic similarity. Formally, if $\mathcal{M}_t = { m_1, m_2, \dots, m_N }$ is the set of memory entries at step $t$, the retrieved set $\hat{\mathcal{M}}_t$ is:

\begin{equation}
\hat{\mathcal{M}}_t = \operatorname{TopK}\Big( \{ m_i \in \mathcal{M}_t \mid \operatorname{sim}(q_{t-1}, m_i) \} \Big)    
\end{equation}

Under this unified formulation, the agent’s observation is given by
\begin{equation}
\label{formula: o_t}
o_t = \{ o_t^{env}, m_t^{scr}, \hat{\mathcal{M}}_t \},
\end{equation}
where $\hat{\mathcal{M}}_t$ denotes selective retrieval, and $m_t^{scr}$ corresponds to deterministic retrieval.

The hybrid memory retrieval ranks information by importance: critical information is maintained in the scratchpad, while potentially useful information is stored in the vector database. Experiments show that this multi-path retrieval setup enhances both the effectiveness and robustness of the agent’s memory.

\subsection{Optimization Strategy}
Since memory operations are realized as structured tokens in the model's vocabulary, optimizing the output sequence likelihood implicitly optimizes the memory policy. We refine the policy using Group Relative Policy Optimization (GRPO)~\citep{GRPO} to master complex memory management in multi-turn scenarios.

During RL, each training sample corresponds to a multi-step trajectory~$\tau = (o_1, a_1, \dots, o_T, a_T)$,
where $a_t$ includes the task-specific action $a_t^{env}$ and the memory operation $a_t^{mem}$,
and the observation $o_t$ is defined in \Cref{formula: o_t}. We use task-level success as the reward signal, i.e., no intermediate rewards are provided and only a terminal reward is assigned at the end of the trajectory. After obtaining the reward, we compute the advantage using the following formulation:
\begin{equation}
    A_i = r_i - \frac{1}{|G|} \sum_{j \in G} r_j
\end{equation}
where $G$ denotes the set of trajectories corresponding to repeated executions of the same task, and $r_i$ is the terminal reward of the i-th trajectory. Following Dr.GRPO~\citep{DRGRPO}, we do not apply normalization to the advantages.

Finally, the advantage is uniformly distributed across all output tokens in the trajectory and optimized according to the following objective:
\begin{equation}
    \mathcal{J}(\theta) = \mathbb{E} \left[ \frac{1}{G} \sum_{i=1}^{G} \rho_{\theta}^{i} A_i - \beta \mathbb{D}_{\text{KL}} [\pi_{\theta} \| \pi_{\text{ref}}] \right]
\end{equation}
Here, $\rho_{\theta}^{i}$ denotes the importance sampling ratio for the $i$-th sample.

Notably, we apply task-level advantages uniformly across all tokens, including memory operations. This enables the agent to jointly optimize memory usage and task performance via RL without external modules.

\begin{table*}[!t]
\centering
\caption{Results on long-context QA benchmarks and multi-turn web benchmarks.
}
\label{tab:model_performance}
\small 
\begin{tabular}{@{}lccccccccc@{}}
\toprule
\multirow{2}{*}{\textbf{Method}} & 
\multicolumn{2}{c}{\textbf{HotpotQA}} & 
\multicolumn{2}{c}{\textbf{2WikiMQA}} & 
\multicolumn{2}{c}{\textbf{Musique}} & \multirow{2}{*}{\textbf{GAIA}} & \multirow{2}{*}{\textbf{WebWalker}} & \multirow{2}{*}{\textbf{Avg.}}\\
\cmidrule(lr){2-3} \cmidrule(lr){4-5} \cmidrule(lr){6-7}
& 200doc & 800doc & 200doc & 800doc & 200doc & 800doc &\\
\midrule
\textit{Training-free Methods}  & & & & & & & & & \\
Full Context  & 63.5 & 62.0 & 55.7 & 49.2 & 42.8 & 41.9 & 23.3 & 29.5 & 46.0\\
Vanilla RAG  & 67.8 & 63.1 & 46.5 & 40.0 & 38.5 & 37.1 & 20.4 & 24.0 & 42.2\\
Generative Agents  & 38.8 & 10.0 & 12.3 & 2.0 & 19.8 & 8.4 & 22.3 & 29.5 & 17.9\\
Mem0 & 38.2 & 33.9 & 24.2 & 18.3 & 14.0 & 11.2 & 25.2 & 28.3 & 24.2\\
A-Mem & 73.5 & 70.4 & 62.7 & 57.1 & 47.1 & 41.6 & 30.1 & 29.0 & 51.4\\
\midrule
\textit{Trained Methods} & & & & & & & & & \\
MemAgent & 76.5 & 71.1 & 65.8  & 57.7 & 54.7 & 44.5 & 33.0 & \textbf{50.0} & 56.7\\
\midrule
\ourwork w/o RL & 65.9 & 60.1 & 52.8 & 55.0 & 47.8 & 40.0 & 35.2 & 45.6 & 50.3\\
\rowcolor{lightyellow}
\textbf{\ourwork~(ours)} & \textbf{77.8} & \textbf{72.9} & \textbf{67.5} & \textbf{62.5} & \textbf{55.1} & \textbf{48.5} & \textbf{37.4} & 48.7 & \textbf{58.8}\\
\bottomrule
\end{tabular}
\end{table*}



\section{Experiments}
In this section, we first introduce our evaluation task and then present the experimental results.
\subsection{Evaluation Tasks}
\label{section:task construction}
\subsubsection{Long Context Benchmarks}
We collect 3 QA datasets: HotpotQA~\citep{hotpotQA}, 2WikiMultiHopQA~\citep{2wiki}, and MuSiQue~\citep{Musique} as our data sources. All three datasets include training and test splits. We perform RL training on the training split and evaluation on the test split.
We feed the document to the model, instructing it to memorize information relevant to the question, and finally require the agent to answer using only the memories.
We augment the difficulty of these QA datasets along the following two dimensions.

\paragraph{Long-context Setting}
Following the RULER~\citep{ruler} benchmark, we construct arbitrary long-context tasks using the following method:
We shuffle the relevant documents and interleave them with a large number of irrelevant documents, constructing a needle-in-a-haystack (NIAH)–style task. This augmentation challenges the agent’s ability to identify and remember important information from massive amounts of input. We train on inputs containing 200 documents (about 28K tokens), and at test time scale the setting to 800 documents (about 112K tokens).

\paragraph{Multi-question Setting}
Following MEM1~\citep{MEM1} and Memory-R1~\citep{Memory-R1}, we provide the model with multiple questions simultaneously. The documents relevant to these questions are shuffled and mixed together before being fed to the agent. After processing all documents, the model is required to answer each question individually. This augmentation strategy challenges the model’s ability to manage and maintain multiple semantically independent memories at the same time. Each task contains a randomly sampled number of questions, ranging from 1 to 10.

\subsubsection{Web Benchmarks}
In multi-turn web search scenarios, we chose the Asearcher~\citep{Aseacher} open-source dataset for training and evaluated our model on the GAIA~\citep{GAIA} and WebWalkerQA~\citep{WebWalker} datasets. In addition to the memory management API, we provide two external tools—Google search engine and Jina URL Reader—to enable the model to access the internet. The maximum number of web tool calls per task is set to $40$ to fully showcase the contribution of the agent’s memory.

\subsection{Baselines}
We evaluate our method against static baselines with hand-designed strategies mem0~\citep{mem0}, A-Mem~\citep{Amem}, GenerativeAgents~\citep{GenerativeAgents}. We also compared our approach with MemAgent~\citep{MemAgent}, another paradigm that employs RL for training Agent Memory. For the implementation of MemAgent, we used the same training hyperparameters and random seed.
Comparisons also include standard RAG and a full-context baseline, with implementation details provided in Appendix \ref{appendix: hyperparameters}.

\subsection{Implementation Details}
\paragraph{Models}
For all agents, we use Qwen3-8B~\citep{Qwen3} as the base model. For agents that require retrieval, we use Qwen3-embedding-0.6B as the embedding model.
\paragraph{Agent Implementation}
We implement memory using a FAISS vector database as the underlying storage. The query for retrieval is provided by the read action. Each action and its XML format will be detailed in Appendix \ref{appendix: hyperparameters}.
Long texts are split into chunks of 4k tokens and fed to the agent step-by-step. At each step, if a read operation is triggered, the memory module retrieves 6 relevant entries from the database.


\paragraph{RL Implementation}
We adopt a fully on-policy RL strategy, where each rollout is used for a single update. For QA tasks, we use exactly match (EM) between the model answer and the ground truth as the reward, and for web tasks, we use an LLM-as-a-judge reward. Additional hyperparameters are provided in the Appendix \ref{appendix: hyperparameters}.

\subsection{Main Results}
The main experimental results are shown in \Cref{tab:model_performance}. 
We highlight the following observations:

(1) \textbf{\ourwork achieves superior performance and robust scalability across varying task scales.} It outperforms all trained and untrained baselines on average. Notably, in the 800-document setting—a 4$\times$ extension of the training context—our model maintains a significant performance lead. This indicates that the agent has learned a content-aware memory policy capable of mitigating information overload as environmental noise increases.

(2) \textbf{RL training substantially optimizes the agent’s memory policy, resulting in large performance gains.} 
After RL training, \ourwork improves by nearly 9 percentage points on average across different task settings. This improvement indicates that directly optimizing memory decisions with task-level feedback is critical for long-horizon tasks. In particular, RL enables the agent to refine when and how memory operations are applied, leading to markedly stronger end performance.

\subsection{Training Dynamic Analysis}
In this section, we provide a detailed analysis of the RL training dynamics of \ourwork.

\begin{figure}[!t]
    \centering
    \includegraphics[width=\linewidth]{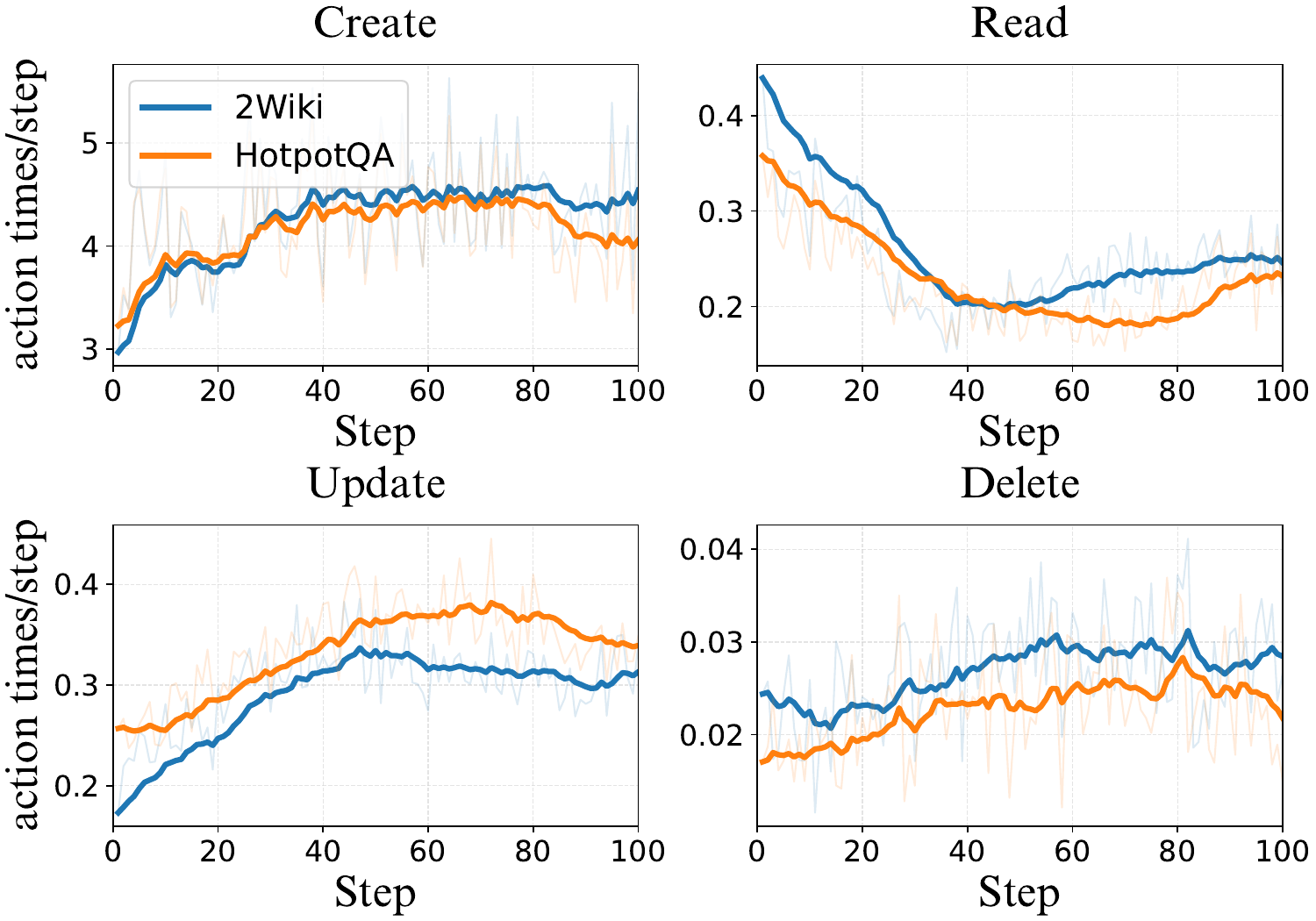}
    \caption{The frequency of the memory operations during the RL training on QA Tasks. The y-axis represents the average number of memory API calls per step made by the model. For instance, a value of 4 indicates that the LLM create 4 memory entries each step.}
    \label{fig:action_rate}
\end{figure}

As shown in \Cref{fig:action_rate}, RL training on QA tasks induces systematic changes in the agent’s memory operation usage.
Specifically, we have the following findings:

(1) \textbf{The model’s behavior shifts from under-managed to task-aligned memory usage.}
Early in training, the model over-relies on Read actions and largely neglects memory maintenance, leading to redundant retrievals. As training progresses, Read usage decreases sharply, while Create, Update, and Delete actions increase substantially. This transition indicates that the model learns to maintain a compact, task-relevant memory by preserving useful information, revising outdated entries, and removing redundancy.

(2) \textbf{While the frequency of Update action remains low compared to Create action, they represent the critical few that significantly influence the agent's overall performance.}
We conduct an ablation study as shown in \Cref{tab:ablation_memory}, demonstrating that removing Update operations leads to a substantial performance drop across all benchmarks, indicating that selectively revising existing memories is critical for maintaining accurate and compact representations as new evidence arrives. In contrast, disabling Delete has only a marginal impact, suggesting that explicit memory removal is less crucial under the current task, which are largely information-accumulation tasks with non-conflicting facts. To verify this, we added experiments in Appendix \ref{appendix: limitation experiment}, showing that when the maximum number of memory entries is limited, the frequencies of these actions exhibit different trends of change, while the importance of actions such as Update and Delete becomes greater.

\begin{table}[t]
\centering
\small
\setlength{\tabcolsep}{3.5pt} 
\caption{Ablation study of memory operations and memory components. Percentage values in brackets represent the relative performance decrease.}
\label{tab:ablation_memory}
\begin{tabular}{llll}
\toprule
\textbf{Method} & \textbf{HotpotQA} & \textbf{2WikiMQA} & \textbf{Musique} \\
\midrule
\textbf{\ourwork}        & 77.8 & 67.5 & 55.1 \\
\midrule
\multicolumn{4}{l}{\emph{Selective Memory Operations}} \\
\quad w/o Update         & 71.4 {\scriptsize (-6.4)} & 62.6 {\scriptsize (-4.9)} & 47.9 {\scriptsize (-7.2)} \\
\quad w/o Delete         & 76.5 {\scriptsize (-1.3)} & 67.3 {\scriptsize (-0.2)} & 54.2 {\scriptsize (-0.9)} \\
\midrule
\multicolumn{4}{l}{\emph{Memory Components}} \\
\quad w/o scratchpad     & 71.8 {\scriptsize (-6.0)} & 56.3 {\scriptsize (-11.2)} & 46.0 {\scriptsize (-9.1)} \\
\quad w/o storage        & 69.2 {\scriptsize (-8.6)} & 59.4 {\scriptsize (-8.1)} & 43.9 {\scriptsize (-11.2)} \\
\quad w/o Both           & 25.6 {\scriptsize (-52.2)} & 27.1 {\scriptsize (-40.4)} & 12.1 {\scriptsize (-43.0)} \\
\bottomrule
\end{tabular}
\end{table}

\begin{figure}[!t]
    \centering
    \includegraphics[width=\linewidth]{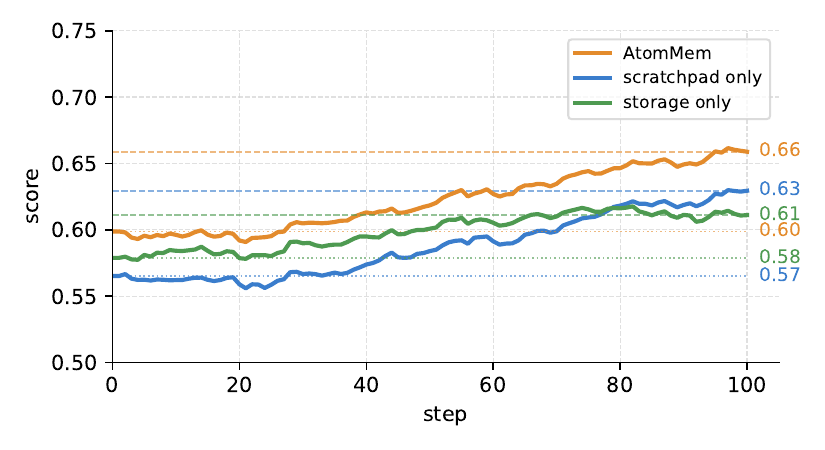}
    \caption{Training curves on HotpotQA for optimizing AtomMem and several of its ablation variants.}
    \label{fig:train_curve}
\end{figure}

\subsection{Ablations}
In this section, we conduct ablation studies on the various memory components of \ourwork and examine the impact of some hyperparameters.
\paragraph{Ablation of Memory Component}
This experiment investigates the contribution of components to the final performance.
The ablation results are reported in \Cref{tab:ablation_memory}. We can see that:

(1) \textbf{\ourwork exhibits robustness to the removal of individual memory components.} Removing either the scratchpad or the external memory storage leads to a moderate performance drop, whereas removing both results in a catastrophic degradation exceeding 40 points. This suggests that when one component is unavailable, the learned policy can still rely on the remaining component to preserve most task-relevant information, rather than collapsing entirely. This indicates that \ourwork is robust to component-level failures.

\begin{figure*}
    \centering
    \includegraphics[width=\linewidth]{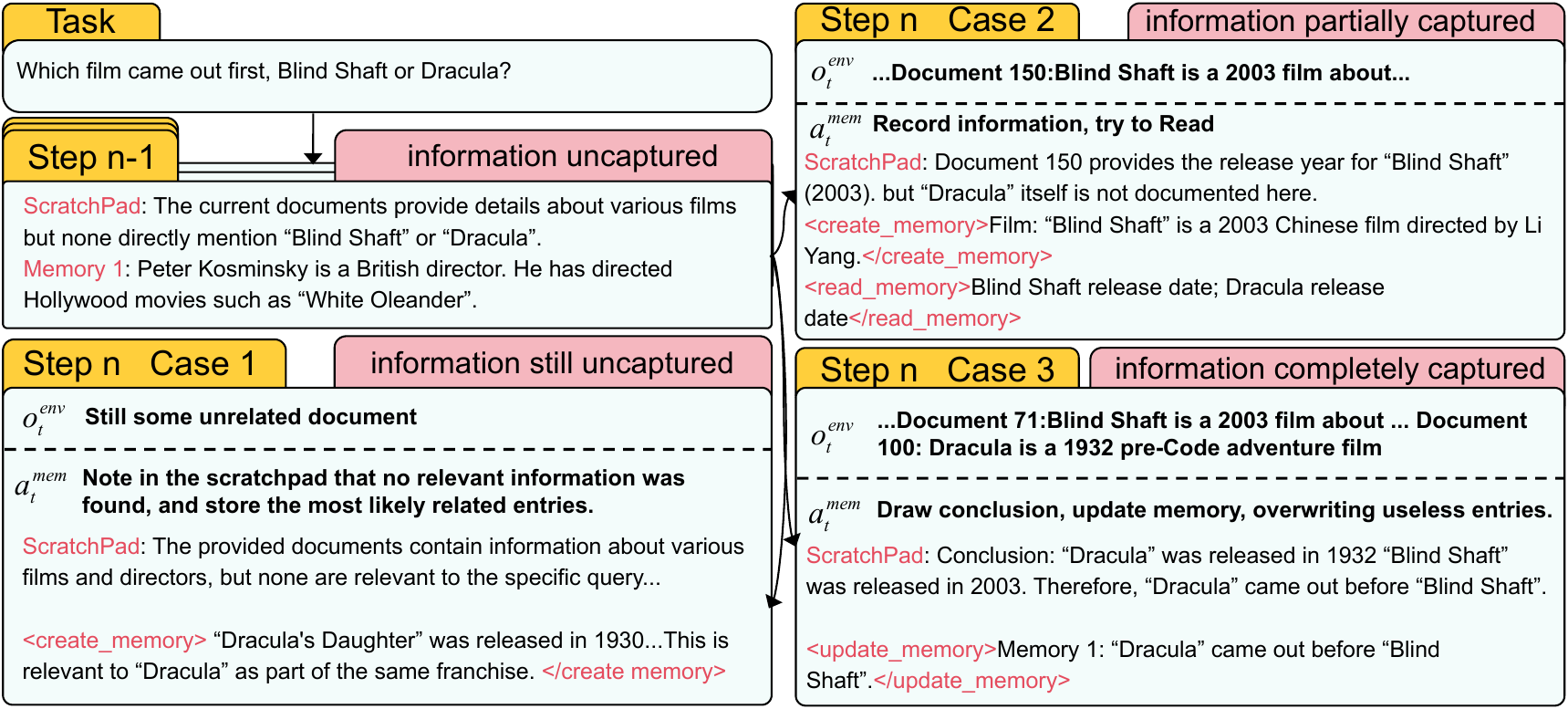}
    \caption{A case illustrates that the model adopts different memory management strategies~($a_t^{mem}$) when facing different task contexts $o_t^{env}$. It demonstrates the dynamic nature of \ourwork.}
    \label{fig:case_study}
\end{figure*}

(2) \textbf{Both the memory storage and the scratchpad contribute substantially to the final performance of \ourwork.} Removing either component leads to a consistent performance drop of 5–10 points across all benchmarks. This indicates that the information preserved by the scratchpad and the external memory storage differ fundamentally in domain and usage, such that neither can be fully substituted by the other.

To verify this, we trained another two variants of \ourwork from scratch: scratchpad-only and storage-only. The results are shown in \Cref{fig:train_curve}. From the experimental results, we observe that \textbf{the other two variants do not achieve performance comparable to \ourwork.} The scratchpad-only variant remains consistently below \ourwork during training, whereas the storage-only variant benefits marginally from RL. This indicates that our design effectively raises the performance ceiling of the agent. The significant gap between \ourwork and its variants suggests that the synergy between the scratchpad and memory storage is a structural necessity for handling complex tasks.

\paragraph{Effect of Hyper-Parameters}
In this experiment, we investigate the effect of several key hyperparameters of \ourwork, including chunk size and retrieve number. The chunk size determines the length of the text segment processed by the agent at each step, while the retrieve number specifies how many entries are retrieved from storage at each step. The experimental results are shown in \Cref{tab:hyperparam_analysis}. From the results, we find that:

(1) \textbf{The retrieval size $K$ must match the task’s memory demand.} Reducing $K$ from 6 to 3 causes a clear performance drop, while increasing it to 12 brings little benefit. This is because the evaluated benchmarks require only 2–4 hop reasoning, for which retrieving about six documents is sufficient.

(2)~\textbf{\ourwork is robust to chunk size.} Performance remains consistent across different chunk sizes, due to the base model’s strong long-context understanding and reinforcement learning that enables effective information extraction at varying granularities.

\paragraph{Sensitivity to Embeddings}
We evaluate the impact of embedding models on task performance. The experimental results are shown in the~\cref{tab:embedding_analysis}. Frow which we find that:

(1)~\textbf{The learned embeddings consistently outperforms random selection across all three tasks.}
Random selection suffers an average drop of 7.4 percentage points compared to Qwen3-embedding-0.6B. This significant gap indicates that similarity-based matching plays an important role in task performance.

(2)~\textbf{The Qwen3 embeddings show a clear trend of improvement as model size increases}. Qwen3-embedding-0.6B achieves an average score of 66.5, while the 4B and 8B variants reach 67.2 and 68.2, respectively. This indicates that higher-capacity embedding models can better capture semantic information relevant to the tasks. Overall, the results highlight the importance of choosing an appropriate embedding model for downstream performance.

\begin{table}[t]
\centering
\small
\setlength{\tabcolsep}{4pt} 
\caption{Hyperparameter Analysis of Chunk Size ($C$) and Retrieve Number ($K$).}
\label{tab:hyperparam_analysis}
\begin{tabular}{
c@{\hspace{6pt}}
c@{\hspace{10pt}}
c@{\hspace{6pt}}
c@{\hspace{6pt}}
c@{\hspace{8pt}}
c
}
\toprule
\textbf{$C$} & \textbf{$K$} & \textbf{HotpotQA} & \textbf{2WikiMQA} & \textbf{Musique} & \textbf{Avg.} \\
\midrule
2048 & 3  & 76.2 & 64.2 & 49.9 & 63.4 \\
2048 & 6  & 76.8 & 67.6 & 52.6 & 65.7 \\
2048 & 12 & 75.0 & 67.3 & 52.6 & 65.0 \\
\midrule
4096 & 3  & 74.2 & 64.8 & 54.6 & 64.5 \\
4096 & 6  & 76.9 & 67.5 & 55.1 & 66.5 \\
4096 & 12 & 77.4 & 69.6 & 54.5 & 67.2 \\
\midrule
8192 & 3  & 76.4 & 65.2 & 53.0 & 64.9 \\
8192 & 6  & 78.7 & 67.9 & 54.0 & 66.9 \\
8192 & 12 & 76.5 & 67.5 & 52.8 & 65.6 \\
\bottomrule
\end{tabular}
\end{table}

\begin{table}[t]
\centering
\small
\setlength{\tabcolsep}{4pt}
\caption{Sensitivity Analysis of Embedding Models.}
\label{tab:embedding_analysis}
\begin{tabular}{
c@{\hspace{10pt}}
c@{\hspace{6pt}}
c@{\hspace{6pt}}
c@{\hspace{8pt}}
c
}
\toprule
\textbf{Embedding Model} & \textbf{HQA} & \textbf{2Wiki} & \textbf{Musi.} & \textbf{Avg.} \\
\midrule
Random Select  & 68.5 & 62.6 & 46.1 & 59.1\\
Qwen3-embedding-0.6B & 76.9 & 67.5 & 55.1 & 66.5\\
Qwen3-embedding-4B  & 77.5 & 68.2 & 56.0 & 67.2\\
Qwen3-embedding-8B  & 78.6 & 69.6 & 56.3 & 68.2\\
\bottomrule
\end{tabular}
\end{table}

\section{Case Study}
In this section, we analyze the model’s responses on a case-by-case basis to understand what memory workflow the model has learned. As illustrated in \Cref{fig:case_study}, we present three scenarios at step $n$ that demonstrate the agent’s learned ability to adapt its memory workflow based on the observation $o_t^{env}$. The example is from HotpotQA, where the LLM made different decisions to complete the task depending on the timing and order in which the key documents appeared.

\noindent $\triangleright$ \textbf{Case 1}: When $o_t^{env}$ contains unrelated documents, the agent uses the scratchpad to log the absence of relevant info and only stores potentially related background entries. 

\noindent $\triangleright$ \textbf{Case 2}: When $o_t^{env}$ provides partial information (e.g., the release date of a single film), the agent commits the newly found evidence to memory and proactively generates a \texttt{<read\_memory>} request to retrieve the missing piece. 

\noindent $\triangleright$ \textbf{Case 3}: In the scenario where all required information is present, the agent synthesizes the retrieved facts within the scratchpad to derive the final answer and uses \texttt{<update\_memory>} to overwrite useless entries with the conclusion.

Together, these cases illustrate that the agent has learned a \textbf{context-sensitive} memory workflow, dynamically deciding when to ignore, retrieve, update, or consolidate memories based on the informational sufficiency of the current observation.

\section{Conclusion}
In this paper, we propose \ourwork, which reframes agentic memory management as a dynamic decision-making problem by deconstructing complex workflows into atomic CRUD operations. By optimizing this learnable decision process, \ourwork moves beyond the limitations of static, ``one-size-fits-all'' memory pipelines. Experimental results and training dynamics demonstrate that this approach enables a task-aligned memory policy.

\section*{Limitation}
Despite its effectiveness, RL optimization is computationally intensive. Training an agent model to convergence requires approximately 2 to 3 days on an 8-GPU cluster. This computational overhead may become a bottleneck when scaling our approach to even longer-horizon or noisier tasks.

In reinforcement learning, task-level advantages are typically evenly distributed across all actions. However, in reality, there should exist more precise methods for assigning advantages, aiming to measure the contribution of each memory entry to the successful completion of a task. We do not explore this direction in the current work for two reasons. First, developing a new RL algorithm for a single downstream task~(agent memory) would be unnecessary, as we have already demonstrated that conventional RL methods can effectively optimize performance. Second, accurately evaluating the value of each memory entry is not a trivial problem, and a clear methodology for doing so remains elusive. We leave this problem for future work.

\section*{Ethical Statement}
All data used in this work are sourced from open-source datasets and do not contain personal or private information. The LLM is used solely for writing and sentence refinement.

\bibliography{latex/custom}

\newpage
\appendix

\begin{table*}[!t]
\centering
\caption{Atomic CRUD Operations for Long-Term Memory Management}
\label{tab:memory_ops}
\small
\begin{tabular}{lll}
\toprule
\textbf{Operation} & \textbf{XML Tag Schema} & \textbf{Functionality} \\
\midrule
\texttt{Create} & \small\texttt{<create\_memory>\{content\}</create\_memory>} & Add new entry to the vector database \\
\texttt{Read}   & \small\texttt{<read\_memory>\{query\}</read\_memory>}     & Retrieve top-$k$ relevant entry \\
\texttt{Update} & \small\texttt{<update\_memory>\{memory id: content\}</update\_memory>} & Modify an existing entry by its identifier \\
\texttt{Delete} & \small\texttt{<delete\_memory>memory id</delete\_memory>}    & Permanently remove an entry \\
\bottomrule
\end{tabular}
\end{table*}



\begin{table*}[!t]
\centering
\small
\caption{Efficiency Comparison: Wall Clock Time, Average Output Token, and Model Calls per Task}
\label{tab:efficiency}
\begin{tabular}{lcccc}
\toprule
\textbf{Method} & \textbf{Wall Clock Time (s/task)} & \textbf{Avg. Tokens} & \textbf{Avg. LLM Calls} & \textbf{Avg. Retrieve Calls} \\
\midrule
\rowcolor{lightyellow}
\ourwork~(ours) & 97.6  & 570.5 & 10.9 & 10.9\\
MemAgent & 49.7  & 264.1 & 8.0 & 0.0\\
Mem0 & 247.8 & 431.7 & 12.9 & 88.6\\
Generative Agents & 416.0 & 494.6 & 375.9  & 418.0\\
A-Mem & 662.4 & 237.7 & 400.0 & 402.0\\
\bottomrule
\end{tabular}
\end{table*}

\begin{table}[!t]
\centering
\small
\caption{Reinforcement Learning Hyperparameters}
\label{tab:rl_hyperparams}
\begin{tabular}{l c}
\toprule
\textbf{Hyperparameter} & \textbf{Value} \\
\midrule
RL algorithm              & GRPO \\
Base model                & Qwen3-8B \\
Batch size                & 16 \\
Rollout Group Size        & 16 \\
Learning rate             & 1e-6 \\
Clip Range High           & 0.28~\citep{DAPO} \\
Clip Range Low           & 0.2 \\
Entropy Loss coefficient       & 0 \\
KL Loss coefficient       & 0~ \citep{RAGEN} \\
Training framework        & Verl~\citep{Verl} \\
Hardware                  & NVIDIA A800 \\
Random Seed               & 42 \\
\bottomrule
\end{tabular}
\end{table}

\section{Implementation Details}
\label{appendix: hyperparameters}
In this section, we list the training hyperparameters, which are shared across all training agents. All training is conducted on NVIDIA A800 GPUs.

\subsection{RL Hyperparameters}
All key RL training hyperparameters are shown in~\Cref{tab:rl_hyperparams}.


\subsection{Agent Implementations}

\subsubsection{Action Space Protocol}
As shown in \Cref{tab:memory_ops}, we define four atomic CRUD operations for long-term memory management, each associated with a structured XML schema and explicit parameters. The \texttt{Create} operation inserts new content as a standalone memory entry into the vector database. \texttt{Read} takes a textual query as input and retrieves the top-$k$ most relevant entries based on vector similarity. \texttt{Update} specifies a unique memory identifier along with revised content, enabling selective modification of existing entries. Finally, \texttt{Delete} removes a memory entry by its identifier, permanently clearing it from storage. Together, these operations provide fine-grained and interpretable control over memory creation, access, refinement, and removal.

\subsubsection{LLM Inference Hyperparameters}
The Qwen3 series recommends using a temperature above 0.6 during inference to avoid repetitive outputs and unstable reasoning; therefore, we set the inference temperature of all agents to 0.7. Meanwhile, top-p is set to 1 and top-k is disabled.

\subsection{Baseline Implementations}
In this work, we use the following baseline:

(1) RAG: Each document is individually stored in the vector database (without chunking). During retrieval, for each question, the question itself is used as the query to retrieve six documents, which are then concatenated and fed to the model for answering.

(2) Full Context: We use YaRN scaling to extend the context of Qwen3-8B to 128K tokens to accommodate the 800-document settings. All questions are input to the model simultaneously, and it is required to answer them sequentially.

(3) mem0 $\&$ Amem $\&$ Generative Agents: We follow the same chunking strategy as \ourwork and use official examples to construct the memory library. During retrieval, we adopt the same strategy as RAG: each question is queried separately, and the retrieved results are concatenated before being fed to the model.

\section{Efficiency Analysis}
In this section, we provide a simple efficiency analysis. The notable differences still demonstrate that \ourwork achieves optimal performance at comparatively high efficiency. The result is shown in \Cref{tab:efficiency}. Analyzing the experimental results, we make the following observations: 

(1)~\textbf{\ourwork and MemAgent achieve higher processing efficiency compared to other agent memory workflows.} This is mainly because the other workflows invoke the LLM multiple times for each input, and this serialized process significantly reduces the efficiency of the memory mechanism, making it nearly unscalable. However, the inference latency of AtomMem is slightly higher than that of MemAgent, due to 1) the increased prompt length caused by the integration of multiple tools, and 2) the addition of a database component, whose retrieval incurs extra latency.

\begin{figure}[!t]
    \centering
    \includegraphics[width=\linewidth]{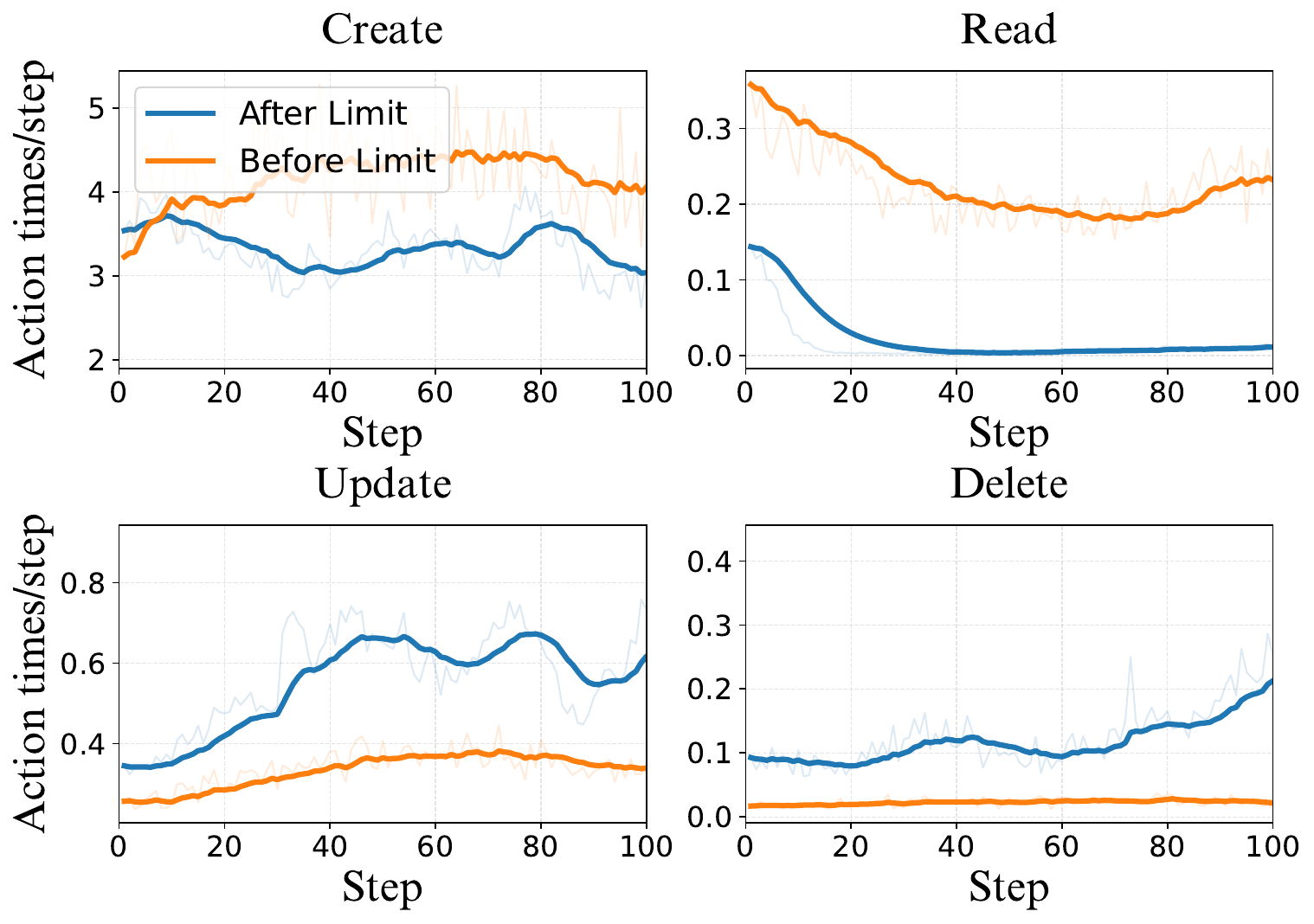}
    \caption{The frequency of memory operations before and after the limitation of database capacity.}
    \label{fig:action_rate_2}
\end{figure}

\section{Memory Capacity Limitation Experiment}
\label{appendix: limitation experiment}
We limit the number of memory entries in the memory store and retrain the agent on the HotpotQA. This allows us to observe changes in action rates, such as whether the importance of the delete action increases. 
Overall, compared with the original experiment, this experiment differs in two aspects: (1) the memory capacity is limited to 20 entries, and any entries exceeding this capacity are discarded; (2) the fine-tuned prompt informs the model of the database limitation and encourages it to make greater use of the Delete operation.
The change in action rates is shown in the \Cref{fig:action_rate_2}.
We observed the following: 

(1)~The frequencies of the Update and Delete operations increased significantly during training, as these operations do not add entries to the memory. This is consistent with our expectations.

(2)~The frequencies of the Create and Read operations decreased significantly. The decrease in Create operations is intuitive, as the model gradually learns to create entries only within the memory capacity. The frequency of Read operations dropped almost to zero, with only a slight rebound near the end of training. We attribute this to that the large number of entries discarded in the early stage. This led to a substantial loss of useful information, causing the model to learn that subsequent Read operations could not retrieve useful content. In the later stages of training, the model learned to retain only useful information in the database, which in turn led to a very slight recovery in the frequency of Read operations.
\section{Prompt}
In this section, we present the prompt structure that remains constant throughout the agent’s execution. The agent’s system prompt and the prompt for its memory fields are shown in \Cref{fig:system_prompt} and \Cref{fig:memory_prompt}.

\begin{figure*}[t]
\centering
\begin{tcolorbox}[
    title=\textbf{\fontsize{8.5}{9.5}\selectfont Agent Prompt},
    colback=TiffanyBlue!5!white, colframe=TiffanyBlue!75!black,
    width=\textwidth,
]
\fontsize{8.5}{10}\selectfont
\textbf{System Prompt:}\\
You are presented with a section of an article and a previous memory. Please learn the provided section carefully and manage your memory to answer questions.

Your short-term memory is a step-wise updated summary, while your long-term memory is a vector database that can be updated through atomic operations. 

short-term memory is updated using <update\_scratchpad>.

Four kinds of memory actions for the database are available:
<read\_memory>: The system maintains a query for memory retrieval. You can modify it via query operations. You will not get any memory unless you give a query. 

<create\_memory>: creates a new entry in the memory. You do not need to repeatedly add the memory shown to you or enter the index of the memory. 

<update\_memory>: updates the existing entry. You need to enter the memory index "Memory i:" to specify which memory to modify. To manage large volumes of memory effectively, you should prioritize using the "modify" function MORE frequently, rather than relying solely on "add" operations.

<delete\_memory>: delete a memory. You need to enter the memory index "Memory i" to specify which memory to delete. You must delete duplicate memory entries!

Use paired XML tags as action markers so you can perform multiple actions—such as adding several memories—in a single response.

action example 1:

<read\_memory>dance partner; Yulia Zagoruychenko.
</read\_memory> 

action example 2:

<create\_memory>

Document 10 indicates that the dance event took place in Moscow in October and that Yulia participated in it. I need to focus more on who else attended this event or who traveled to Moscow in October, in order to infer who Yulia’s dance partner might be.

<create\_memory>

action example 3:

<update\_memory>

Memory 1: The current article provides updated competition records showing that Riccardo Cocchi is now partnered with Emily in the 2025 season, while no recent evidence confirms his continued partnership with Yulia Zagoruychenko. This conflicts with the previous memory stating that Yulia’s partner is Riccardo. Since the information clearly supersedes the earlier record, the correct action is to modify the existing memory to reflect that Yulia’s current dance partner is unknown as of 2025, and mark the entry for re-verification.

</update\_memory>

action example 4:

It can be observed that Entry 2 and Entry 6 are largely duplicated. Since Entry 6 is more recent, I choose to delete Entry 2. 
<delete\_memory>

Memory 2

</delete\_memory>

Response example:

<update\_scratchpad>

\textbf{1. Memory Framework Overview}  

The provided section describes a \textbf{memory-augmented reasoning framework} in which the agent maintains two complementary memory systems. \textbf{Short-term memory} functions as an evolving, step-wise summary that tracks the most immediate context, while \textbf{long-term memory} resides in a vector database capable of being updated through atomic operations.

\textbf{2. Memory Operations}  

The text further introduces \textbf{four atomic memory actions}—\textbf{read\_memory}, \textbf{create\_memory}, \textbf{update\_memory}, and \textbf{delete\_memory}.  
These collectively define how the agent:
- retrieves information through query management,
- adds new knowledge,
- updates outdated or inaccurate entries, and
- removes redundant or conflicting information.

\textbf{3. Functional Role of the Two Memory Types}  
The mechanisms emphasize the importance of \textbf{active and continuous memory management}. Short-term memory captures high-level situational summaries that guide immediate reasoning, while long-term memory stores \textbf{fine-grained, reusable knowledge} that can support future inference, compensate for missing context, and improve the agent’s overall consistency across steps.
</update\_scratchpad>

<read\_memory>

short-term vs long-term memory mechanism; atomic operations; memory workflow

</read\_memory>

<create\_memory>

Long-term memory note: The system's memory architecture explicitly separates short-term and long-term roles. Short-term memory is updated every step and acts as a compressed running summary of what the agent has just read, inferred, or decided. In contrast, the long-term memory is a vector-database-backed store meant to hold more detailed, fine-grained, and broadly relevant information—such as definitions, recurring concepts, protocol rules, and any knowledge that may be useful across multiple future queries. The long-term store is maintained through atomic operations (query updating, adding new facts, modifying older entries, and deleting redundant ones), making it flexible and continually improvable as new context appears.

</create\_memory>
\end{tcolorbox}
\caption{System prompt for the task.}
\label{fig:system_prompt}
\end{figure*}

\begin{figure*}
\centering
\begin{tcolorbox}[
    title=\textbf{\fontsize{8.5}{9.5}\selectfont Agent Prompt},
    colback=TiffanyBlue!5!white, colframe=TiffanyBlue!75!black,
    width=\textwidth,
]
\tcbline
\textbf{User:}
This is the question you need to solve:

{prompt}

This is your scratchpad from the previous turn.

\{short\_memory\}

This is the current query to retrieve memory from the database:

\{query\}

This is the current memory related to the query:

\{long\_memory\}

Tips:
DO NOT repeatedly update the query. If you don’t have the desired memory, it means the entry does not exist in the knowledge base. AVOID using update\_query multiple times within a single response; instead, you can use a long and composite query to retrieve documents for different questions. The query matches documents based on semantic embeddings, and composite queries are best composed of keywords.

This is the article:

\{chunk\}
\end{tcolorbox}
\caption{memory prompt for the task.}
\label{fig:memory_prompt}
\end{figure*}

\end{document}